\PassOptionsToPackage{table}{xcolor}
\documentclass{article}
\usepackage{iclr2027_conference,times}
\usepackage{xcolor}

\usepackage{amsmath,amssymb,amsfonts}
\usepackage{booktabs}
\usepackage{graphicx}
\usepackage{multirow}
\usepackage{placeins}
\usepackage{tikz}
\usepackage{pgfplots}
\usetikzlibrary{arrows.meta,calc}
\pgfplotsset{compat=1.18}
\usepackage{microtype}
\usepackage{hyperref}
\usepackage{url}
\usepackage{float}

\AtBeginDocument{%
  \setlength{\abovedisplayskip}{6pt plus 2pt minus 2pt}%
  \setlength{\belowdisplayskip}{6pt plus 2pt minus 2pt}%
  \setlength{\abovedisplayshortskip}{3pt plus 2pt minus 1pt}%
  \setlength{\belowdisplayshortskip}{4pt plus 2pt minus 2pt}}
\AddToHook{env/figure/begin}{%
  \setlength{\abovecaptionskip}{3pt}\setlength{\belowcaptionskip}{0pt}}
\AddToHook{env/table/begin}{%
  \setlength{\abovecaptionskip}{0pt}\setlength{\belowcaptionskip}{5pt}%
  \exhyphenpenalty=10000}

\newcommand{\bx}{\mathbf{x}}
\newcommand{\bu}{\mathbf{u}}
\newcommand{\bz}{\mathbf{z}}

\newcommand{\R}{\mathbb{R}}
\newcommand{\N}{\mathbb{N}}
\newcommand{\E}{\mathbb{E}}
\newcommand{\Cin}{C_{\mathrm{in}}}
\newcommand{\Cout}{C_{\mathrm{out}}}

\newcommand{\ours}{RBF-GNN}

\newcommand{\thetaglyph}[2]{%
  \begin{scope}[shift={#1}]
    \foreach \cx/\cy/\shd in {%
      0/0/85,1/0/35,2/0/60,0/1/25,1/1/70,2/1/45,0/2/90,1/2/30,2/2/55}
      \fill[#2, fill opacity={\shd*0.01}]
        ({-1.95mm+\cx*1.3mm},{-1.95mm+\cy*1.3mm}) rectangle +(1.15mm,1.15mm);
    \draw[black!45, line width=0.3pt] (-2.05mm,-2.05mm) rectangle (1.95mm,1.95mm);
  \end{scope}}

\def\redc{\cellcolor[HTML]{FF999A}}
\def\orangec{\cellcolor[HTML]{FFCC99}}
\def\yellowc{\cellcolor[HTML]{FFF8AD}}
\newcommand{\tightbox}[2]{%
  {\setlength{\fboxsep}{1pt}\colorbox[HTML]{#1}{\strut #2}}}

\newcommand{\ranklegend}{%
  \tightbox{FF999A}{best},
  \tightbox{FFCC99}{second}, and
  \tightbox{FFF8AD}{third}}

\title{RBF-GNN: Rational Basis Functions for\\Pseudo-Coordinate based Graph Convolutions}

\author{\textbf{Paweł Batorski}\textsuperscript{1,*} \quad
\textbf{Abtin Pourhadi}\textsuperscript{1,*} \quad\textbf{Paul Swoboda}\textsuperscript{1} \\ 
\textsuperscript{1}Heinrich Heine University Düsseldorf \\\textsuperscript{*}Equal contribution \\}

\iclrfinalcopy

\begin{document}

\maketitle
\fancyhead{}
\renewcommand{\headrulewidth}{0pt}

\begin{abstract}
We propose RBF-GNN, a new pseudo-coordinate based graph neural network
architecture that takes into account Euclidean, spherical or angular
coordinates and uses them to induce a powerful spatial inductive bias.
Similar in architecture to SplineCNN, we improve upon the latter by replacing
the less efficient sparse-activation based B-splines whose number grows
exponentially with dimension by rational Pad\'e basis functions. For effective
training we propose a spline-subspace initialization and a variance-preserving
weight rescaling. Experimentally, we evaluate on a number of popular neural
network architectures that use SplineCNNs. We replace only the SplineCNNs with
RBF-GNN. We achieve improved results, including on semantic keypoint matching,
shape matching, event based camera computer vision tasks. Code is available at \url{https://github.com/pawelswoboda/RationalBasisCNN}.
\end{abstract}

\begin{figure}[H]
\centering
\begin{tikzpicture}[
  declare function={
    hp(\x,\y,\cx,\cy)=max(0,1-4*abs(\x-\cx))*max(0,1-4*abs(\y-\cy));
    ratA(\x,\y)=1/((1+3*(2*\x-1.2)^2)*(1+3*(2*\y-0.8)^2));
    ratB(\x,\y)=0.8/((1+2*(2*\x-0.6)^2)*(1+2*(2*\y-1.4)^2));
    ratC(\x,\y)=0.55/((1+3*(2*\x-1.6)^2)*(1+3*(2*\y-1.6)^2));
  },
  gnode/.style={circle, fill=black!75, inner sep=1.4pt},
  cnode/.style={circle, fill=orange!90!black, inner sep=1.8pt},
  gedge/.style={black!60, thin},
  hedge/.style={orange!85!black, line width=0.9pt},
  stem/.style={densely dashed, black!70, thin},
  conn/.style={line width=0.35pt},
]
\begin{axis}[
  name=spl,
  width=0.48\linewidth, height=5.6cm,
  view={25}{30},
  xmin=0, xmax=1, ymin=0, ymax=1, zmin=0, zmax=1.05,
  hide axis,
  clip=false,
  offhat/.style={surf, shader=flat, draw=black!45, fill=black!20,
                 fill opacity=0.15, draw opacity=0.4, line width=0.2pt},
]
\addplot3[fill=black!6, draw=black!40, thin] coordinates
  {(0,0,0) (1,0,0) (1,1,0) (0,1,0) (0,0,0)};
\addplot3[black!25, thin] coordinates {(0.25,0,0) (0.25,1,0)};
\addplot3[black!25, thin] coordinates {(0.5,0,0) (0.5,1,0)};
\addplot3[black!25, thin] coordinates {(0.75,0,0) (0.75,1,0)};
\addplot3[black!25, thin] coordinates {(0,0.25,0) (1,0.25,0)};
\addplot3[black!25, thin] coordinates {(0,0.5,0) (1,0.5,0)};
\addplot3[black!25, thin] coordinates {(0,0.75,0) (1,0.75,0)};
\addplot3[offhat, domain=0:0.5, y domain=0.75:1, samples=3, samples y=2] {hp(x,y,0.25,1)};
\addplot3[offhat, domain=0.5:1, y domain=0.75:1, samples=3, samples y=2] {hp(x,y,0.75,1)};
\addplot3[offhat, domain=0.25:0.75, y domain=0.75:1, samples=3, samples y=2] {hp(x,y,0.5,1)};
\addplot3[offhat, domain=0.75:1, y domain=0.5:1, samples=2, samples y=3] {hp(x,y,1,0.75)};
\addplot3[offhat, domain=0:0.5, y domain=0.5:1, samples=3, samples y=3] {hp(x,y,0.25,0.75)};
\addplot3[offhat, domain=0.75:1, y domain=0.25:0.75, samples=2, samples y=3] {hp(x,y,1,0.5)};
\addplot3[offhat, domain=0:0.5, y domain=0.25:0.75, samples=3, samples y=3] {hp(x,y,0.25,0.5)};
\addplot3[offhat, domain=0.5:1, y domain=0:0.5, samples=3, samples y=3] {hp(x,y,0.75,0.25)};
\addplot3[offhat, domain=0.25:0.75, y domain=0:0.5, samples=3, samples y=3] {hp(x,y,0.5,0.25)};
\addplot3[offhat, domain=0:0.5, y domain=0:0.5, samples=3, samples y=3] {hp(x,y,0.25,0.25)};
\draw[gedge] (axis cs:0.5,0.5,0) -- (axis cs:0.18,0.28,0);
\draw[gedge] (axis cs:0.5,0.5,0) -- (axis cs:0.82,0.30,0);
\draw[gedge] (axis cs:0.5,0.5,0) -- (axis cs:0.30,0.82,0);
\draw[gedge] (axis cs:0.5,0.5,0) -- (axis cs:0.88,0.62,0);
\node[gnode] at (axis cs:0.18,0.28,0) {};
\node[gnode] at (axis cs:0.82,0.30,0) {};
\node[gnode] at (axis cs:0.30,0.82,0) {};
\node[gnode] at (axis cs:0.88,0.62,0) {};
\addplot3[surf, shader=interp, opacity=0.55,
  colormap={hD}{color=(brown!8) color=(brown!70)},
  domain=0.5:1, y domain=0.5:1, samples=17] {hp(x,y,0.75,0.75)};
\addplot3[surf, shader=interp, opacity=0.55,
  colormap={hC}{color=(olive!8) color=(olive!70)},
  domain=0.25:0.75, y domain=0.5:1, samples=17] {hp(x,y,0.5,0.75)};
\addplot3[surf, shader=interp, opacity=0.55,
  colormap={hB}{color=(orange!8) color=(orange!80!black)},
  domain=0.5:1, y domain=0.25:0.75, samples=17] {hp(x,y,0.75,0.5)};
\addplot3[surf, shader=interp, opacity=0.55,
  colormap={hA}{color=(red!8) color=(red!65)},
  domain=0.25:0.75, y domain=0.25:0.75, samples=17] {hp(x,y,0.5,0.5)};
\addplot3[brown!70, thin, opacity=0.9] coordinates
  {(0.5,0.75,0) (0.75,0.75,1) (1,0.75,0)};
\addplot3[brown!70, thin, opacity=0.9] coordinates
  {(0.75,0.5,0) (0.75,0.75,1) (0.75,1,0)};
\addplot3[olive!70, thin, opacity=0.9] coordinates
  {(0.25,0.75,0) (0.5,0.75,1) (0.75,0.75,0)};
\addplot3[olive!70, thin, opacity=0.9] coordinates
  {(0.5,0.5,0) (0.5,0.75,1) (0.5,1,0)};
\addplot3[orange!80!black, thin, opacity=0.9] coordinates
  {(0.5,0.5,0) (0.75,0.5,1) (1,0.5,0)};
\addplot3[orange!80!black, thin, opacity=0.9] coordinates
  {(0.75,0.25,0) (0.75,0.5,1) (0.75,0.75,0)};
\addplot3[red!65, thin, opacity=0.9] coordinates
  {(0.25,0.5,0) (0.5,0.5,1) (0.75,0.5,0)};
\addplot3[red!65, thin, opacity=0.9] coordinates
  {(0.5,0.25,0) (0.5,0.5,1) (0.5,0.75,0)};
\draw[hedge] (axis cs:0.5,0.5,0) -- (axis cs:0.63,0.58,0);
\node[cnode] at (axis cs:0.5,0.5,0) {};
\node[gnode] at (axis cs:0.63,0.58,0) {};
\node[font=\scriptsize, orange!80!black, anchor=north east, inner sep=1pt]
  at (axis cs:0.5,0.5,0) {$i$};
\node[font=\scriptsize, anchor=north west, inner sep=1pt]
  at (axis cs:0.63,0.58,0) {$j$};
\draw[stem] (axis cs:0.63,0.58,0) -- (axis cs:0.63,0.58,0.3536);
\fill[red!65]          (axis cs:0.63,0.58,0.3264) circle (1.1pt);
\fill[orange!80!black] (axis cs:0.63,0.58,0.3536) circle (1.1pt);
\fill[olive!70]        (axis cs:0.63,0.58,0.1536) circle (1.1pt);
\fill[brown!70]        (axis cs:0.63,0.58,0.1664) circle (1.1pt);
\coordinate (pkA) at (axis cs:0.5,0.5,1);
\coordinate (pkB) at (axis cs:0.75,0.5,1);
\coordinate (pkC) at (axis cs:0.5,0.75,1);
\coordinate (pkD) at (axis cs:0.75,0.75,1);
\coordinate (pkE) at (axis cs:1,0.5,1);
\coordinate (pkF) at (axis cs:0.75,1,1);
\coordinate (pkG) at (axis cs:1,0.75,1);
\end{axis}
\begin{axis}[
  at={(spl.outer east)}, anchor=outer west, xshift=6mm,
  name=rat,
  width=0.48\linewidth, height=5.6cm,
  view={25}{30},
  xmin=0, xmax=1, ymin=0, ymax=1, zmin=0, zmax=1.05,
  hide axis,
  clip=false,
]
\addplot3[fill=black!6, draw=black!40, thin] coordinates
  {(0,0,0) (1,0,0) (1,1,0) (0,1,0) (0,0,0)};
\draw[gedge] (axis cs:0.5,0.5,0) -- (axis cs:0.18,0.28,0);
\draw[gedge] (axis cs:0.5,0.5,0) -- (axis cs:0.82,0.30,0);
\draw[gedge] (axis cs:0.5,0.5,0) -- (axis cs:0.30,0.82,0);
\draw[gedge] (axis cs:0.5,0.5,0) -- (axis cs:0.88,0.62,0);
\node[gnode] at (axis cs:0.18,0.28,0) {};
\node[gnode] at (axis cs:0.82,0.30,0) {};
\node[gnode] at (axis cs:0.30,0.82,0) {};
\node[gnode] at (axis cs:0.88,0.62,0) {};
\addplot3[surf, shader=interp, opacity=0.45,
  colormap={rC}{color=(violet!6) color=(violet!65)},
  domain=0:1, y domain=0:1, samples=25] {ratC(x,y)};
\addplot3[surf, shader=interp, opacity=0.45,
  colormap={rB}{color=(teal!6) color=(teal!65)},
  domain=0:1, y domain=0:1, samples=25] {ratB(x,y)};
\addplot3[surf, shader=interp, opacity=0.45,
  colormap={rA}{color=(blue!6) color=(blue!70!black)},
  domain=0:1, y domain=0:1, samples=25] {ratA(x,y)};
\draw[hedge] (axis cs:0.5,0.5,0) -- (axis cs:0.63,0.58,0);
\node[cnode] at (axis cs:0.5,0.5,0) {};
\node[gnode] at (axis cs:0.63,0.58,0) {};
\node[font=\scriptsize, orange!80!black, anchor=north east, inner sep=1pt]
  at (axis cs:0.5,0.5,0) {$i$};
\node[font=\scriptsize, anchor=north west, inner sep=1pt]
  at (axis cs:0.63,0.58,0) {$j$};
\draw[stem] (axis cs:0.18,0.28,0) -- (axis cs:0.18,0.28,0.2975);
\fill[blue!70!black] (axis cs:0.18,0.28,0.2736) circle (0.9pt);
\fill[teal!65]       (axis cs:0.18,0.28,0.2975) circle (0.9pt);
\fill[violet!65]     (axis cs:0.18,0.28,0.0231) circle (0.9pt);
\draw[stem] (axis cs:0.82,0.30,0) -- (axis cs:0.82,0.30,0.5648);
\fill[blue!70!black] (axis cs:0.82,0.30,0.5648) circle (0.9pt);
\fill[teal!65]       (axis cs:0.82,0.30,0.1109) circle (0.9pt);
\fill[violet!65]     (axis cs:0.82,0.30,0.1368) circle (0.9pt);
\draw[stem] (axis cs:0.63,0.58,0) -- (axis cs:0.63,0.58,0.7124);
\fill[blue!70!black] (axis cs:0.63,0.58,0.7124) circle (1.1pt);
\fill[teal!65]       (axis cs:0.63,0.58,0.3834) circle (1.1pt);
\fill[violet!65]     (axis cs:0.63,0.58,0.2583) circle (1.1pt);
\draw[stem] (axis cs:0.30,0.82,0) -- (axis cs:0.30,0.82,0.7174);
\fill[blue!70!black] (axis cs:0.30,0.82,0.1543) circle (0.9pt);
\fill[teal!65]       (axis cs:0.30,0.82,0.7174) circle (0.9pt);
\fill[violet!65]     (axis cs:0.30,0.82,0.1368) circle (0.9pt);
\draw[stem] (axis cs:0.88,0.62,0) -- (axis cs:0.88,0.62,0.3678);
\fill[blue!70!black] (axis cs:0.88,0.62,0.3259) circle (0.9pt);
\fill[teal!65]       (axis cs:0.88,0.62,0.2062) circle (0.9pt);
\fill[violet!65]     (axis cs:0.88,0.62,0.3678) circle (0.9pt);
\coordinate (qkA) at (axis cs:0.6,0.4,1);
\coordinate (qkB) at (axis cs:0.3,0.7,0.8);
\coordinate (qkC) at (axis cs:0.8,0.8,0.55);
\end{axis}
\coordinate (lrow) at ([yshift=8mm]spl.north);
\node[font=\scriptsize] at ([yshift=5mm]lrow)
  {$k^D$ weight matrices $\Theta_p$ --- one per spline};
\thetaglyph{([xshift=-21mm]lrow)}{red!65}
\thetaglyph{([xshift=-15mm]lrow)}{orange!80!black}
\thetaglyph{([xshift=-9mm]lrow)}{olive!70}
\thetaglyph{([xshift=-3mm]lrow)}{brown!70}
\thetaglyph{([xshift=3mm]lrow)}{black!30}
\thetaglyph{([xshift=9mm]lrow)}{black!30}
\thetaglyph{([xshift=15mm]lrow)}{black!30}
\node at ([xshift=21mm]lrow) {$\cdots$};
\draw[conn, red!60]          ([xshift=-21mm,yshift=-2.2mm]lrow) -- (pkA);
\draw[conn, orange!70!black] ([xshift=-15mm,yshift=-2.2mm]lrow) -- (pkB);
\draw[conn, olive!60]        ([xshift=-9mm,yshift=-2.2mm]lrow)  -- (pkC);
\draw[conn, brown!60]        ([xshift=-3mm,yshift=-2.2mm]lrow)  -- (pkD);
\draw[conn, black!25]        ([xshift=3mm,yshift=-2.2mm]lrow)   -- (pkE);
\draw[conn, black!25]        ([xshift=9mm,yshift=-2.2mm]lrow)   -- (pkF);
\draw[conn, black!25]        ([xshift=15mm,yshift=-2.2mm]lrow)  -- (pkG);
\coordinate (rrow) at (lrow -| rat.north);
\node[font=\scriptsize] at ([yshift=5mm]rrow)
  {$K$ weight matrices $\Theta_p$ --- one per basis function};
\thetaglyph{([xshift=-8mm]rrow)}{blue!70!black}
\thetaglyph{(rrow)}{teal!65}
\thetaglyph{([xshift=8mm]rrow)}{violet!65}
\draw[conn, blue!60!black] ([xshift=-8mm,yshift=-2.2mm]rrow) -- (qkA);
\draw[conn, teal!60]       ([yshift=-2.2mm]rrow)             -- (qkB);
\draw[conn, violet!60]     ([xshift=8mm,yshift=-2.2mm]rrow)  -- (qkC);
\node[font=\small] at ([yshift=-4mm]spl.south)
  {(a) SplineCNN: B-spline basis};
\node[font=\small] at ([yshift=-4mm]rat.south)
  {(b) \ours: rational basis};
\end{tikzpicture}
\caption{\ours{} replaces the fixed B-spline basis of SplineCNN by a small
set of global, learnable rational basis functions. Both panels show the
pseudo-coordinate domain $[0,1]^D$, $D=2$ with the neighborhood of a node $i$: each
neighbor $j$ sits at its pseudo-coordinate $\bu_{ij}$. Every basis function
carries one weight matrix $\Theta_p$ (small squares), and the dashed stem
over the highlighted edge $(i,j)$ marks the basis values that weight its
message. \textbf{(a)}~SplineCNN has one weight matrix for each of the $k^D$ B-splines on the knot grid of dimension $D$, but an edge activates only the $2^D$ splines
around its pseudo-coordinate (colored); all other splines (faint) and their matrices (gray) receive no signal from this edge. \textbf{(b)}~\ours{} uses $K$ rational
safe Pad\'e basis functions (here $K=3$), smooth, globally supported, learnable, with dense activations and $K$ is independent of $D$.}
\label{fig:teaser}
\end{figure}
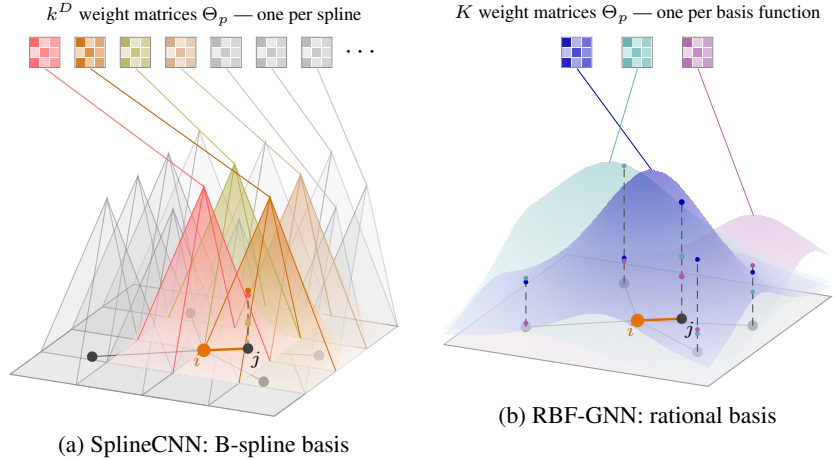

\section{Introduction}
\label{sec:intro}

Many graph convolutions for geometric data define their kernel as a continuous
function of pseudo-coordinates attached to the edges, and they differ mainly in
how this function is represented. SplineCNN~\citep{fey2018splinecnn} expands it
in a fixed tensor-product basis of B-splines, MoNet~\citep{monti2017monet} in a
mixture of learnable Gaussians, and more recent continuous-kernel
operators~\citep{ma2024ckgconv} in a small multilayer perceptron of the
pseudo-coordinates. This choice determines which kernels a layer can represent
and how many parameters it requires.

SplineCNN has been especially powerful and has been adopted in a range of later architectures. Its basis functions are compactly supported B-splines on a regular grid over the unit cube. This gives the kernel an explicit spatial locality and makes the evaluation time independent of the kernel size, because only the few basis functions whose support contains the pseudo-coordinate of an edge have to be evaluated.
The same grid, however, also determines the number of parameters. With $k$ basis functions per axis in $D$ pseudo-coordinate dimensions, the grid contains $K = k^D$ basis functions, and the layer has one weight matrix for each of them. For a fixed resolution per axis, the number of weight matrices therefore grows exponentially with $D$, independently of the complexity of the kernel that has to be learned. Because of the compact support, each edge moreover involves only a small subset of these weight matrices, so that each weight matrix is updated by only a fraction of the edges.

In this work, we replace the B-spline basis by $K$ learnable rational functions, so that $K$ becomes a hyperparameter that is independent of $D$. Each basis function is a ratio of two polynomials in the safe Pad\'e form $B = P(t)/(1 + |Q(t)|)$~\citep{molina2020pade}, whose denominator is at least one and therefore never vanishes, and both polynomials are expressed in the Chebyshev basis for better numerical conditioning. Unlike the B-splines, these functions are not compactly supported, so every edge involves all $K$ weight matrices. Since rational functions with small random coefficients are nearly flat, we initialize the basis from the B-splines it replaces. When $K = k^D$, we fit the B-splines directly. Otherwise, we fit their leading principal components and rescale the kernel weights so that the initial message variance matches that of the B-spline layer.

In our experiments, we obtain performance improvements whenever we replace SplineCNN with \ours{} across a wide array of applications and deep learning architectures.
Additionally, in many applications we can do so with significantly lower parameter counts.

In summary, our contributions are as follows:

\begin{itemize}
    \setlength{\itemsep}{2pt}
    \setlength{\topsep}{4pt}
    \setlength{\parsep}{0pt}
    \item We propose \ours{}, a new pseudo-coordinate based GNN architecture with learnable activation functions. Our formulation is efficient and does not suffer from a blow-up in the number of activations.
\item We propose practical ways to make \ours{} work, namely a special initialization scheme and a fast kernel implementation.
\item  We evaluate  \ours{} across eight benchmarks whose
pseudo-coordinate dimension ranges from two to six, spanning keypoint matching
on PascalVOC~\citep{everingham2010pascal} and SPair-71k~\citep{min2019spair}
in two published matching architectures~\citep{fey2020dgmc,normmatchtrans},
shape correspondence on FAUST~\citep{bogo2014faust}, event-camera recognition
on N-Caltech101~\citep{orchard2015converting} and N-Cars~\citep{sironi2018hats},
superpixel classification~\citep{dwivedi2023benchmarking} and point-cloud part
segmentation~\citep{yi2016scalable}.
We get consistent performance improvements by switching out our closest competitor SplineCNN to \ours{} with a matched or sometimes even lower parameter count and similar or lower compute costs.
\end{itemize}

\section{Related work}
\label{sec:related}

\paragraph{Geometry-aware graph convolutions.}
Message-passing neural networks provide the general neighbourhood-aggregation
view~\citep{gilmer2017mpnn}, while spatial operators additionally condition
messages on edge geometry. ECC generates filters from edge attributes
\citep{simonovsky2017ecc}; MoNet applies Gaussian mixtures to
pseudo-coordinates~\citep{monti2017monet}; and FeaStNet learns
feature-dependent assignments to weight matrices~\citep{verma2018feastnet}.
Continuous filters have also been generated by neural networks for molecules
\citep{schutt2017schnet}, point clouds~\citep{wang2018pccnn,wu2019pointconv},
and general graphs equipped with positional encodings~\citep{ma2024ckgconv}.
Related point-cloud operators use Taylor-polynomial filters
\citep{xu2018spidercnn}, learnable kernel points~\citep{thomas2019kpconv},
geometry-conditioned weight banks~\citep{xu2021paconv}, or relative-position
attention~\citep{zhao2021pointtransformer}. SplineCNN instead expands a
continuous kernel in compactly supported tensor-product B-splines
\citep{fey2018splinecnn}, an operator subsequently used for graph matching
\citep{fey2020dgmc} and event streams~\citep{schaefer2022aegnn}. RBF-GNN is
closest to SplineCNN: it preserves its message-passing rule and matrix bank,
but replaces the fixed grid basis by learnable non-separable rational
functions whose number is independent of $k^D$.

\paragraph{Learned continuous and rational bases.}
Learnable functional parameterisations include spline time--frequency atoms
\citep{balestriero2018splinefilters}, neural continuous kernels
\citep{romero2022ckconv,romero2022flexconv}, and coordinate networks based on
Fourier features, periodic activations, or multiplicative filters
\citep{tancik2020fourier,sitzmann2020siren,sahu2021mfn}. Rational functions
have been studied as trainable activations~\citep{molina2020pade} and as
efficient smooth approximators~\citep{boulle2020rational}. More recently, KAN
places learned B-spline functions on network edges~\citep{liu2025kan}, while
KAT replaces those splines by rational functions to improve efficiency and
scalability~\citep{yang2025kat}. Rational-activation and KAN-family models
modify neuron or feature-mixing nonlinearities, while neural coordinate models
learn an entire continuous signal or kernel. In contrast, our rational
functions are multivariate mixing coefficients of a matrix-valued graph kernel
and are initialised to preserve both the SplineCNN kernel subspace and its
output variance.

\section{Method}
\label{sec:operator}

\begin{figure}[t]
\centering
\includegraphics[width=0.96\linewidth]{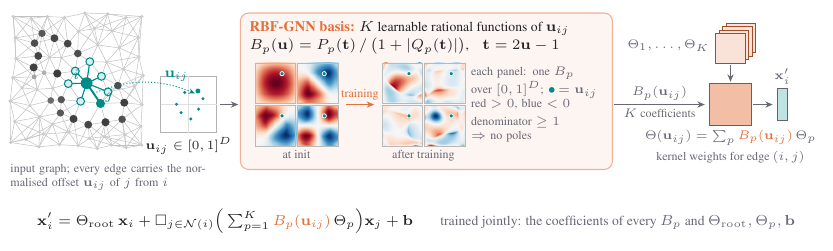}
\caption{Overview of \ours{}. Each graph edge $(j,i)$ carries a normalized pseudo-coordinate $\bu_{ij}$. The $K$ learnable safe rational basis functions
are evaluated at $\bu_{ij}$ and mix the weight matrices $\Theta_p$ into the
edge-specific continuous kernel $\Theta(\bu_{ij})$. The resulting neighbor
messages are aggregated to produce $\bx'_i$; the rational-basis coefficients
and convolution weights are trained jointly.}
\label{fig:method}
\end{figure}

\ours{} is a graph convolution whose continuous kernel is spanned by learnable rational basis functions of pseudo-coordinates.
We first set up pseudo-coordinates and the continuous-kernel convolution operator, then introduce the rational basis in safe Pad\'e form.
We relate our construction to SplineCNN~\citep{fey2018splinecnn}.
Finally, we assemble the full network and describe the initialization that lets it train as reliably as the spline model it can replace. Figure~\ref{fig:method} summarizes the complete pipeline.

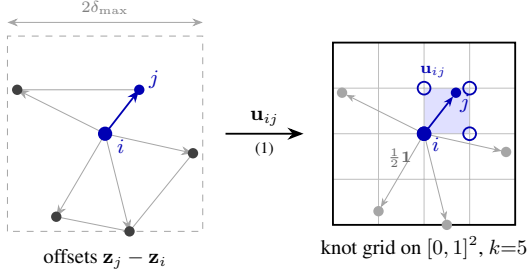
\begin{figure}[t]
\begin{minipage}[c]{0.50\linewidth}
\centering
\resizebox{\linewidth}{!}{%
\begin{tikzpicture}[
  every node/.style={font=\small},
  vertex/.style={circle,fill=black!75,inner sep=1.6pt},
]
\begin{scope}[scale=1.5]
\coordinate (i)  at (0,0);
\coordinate (j1) at (-0.90,0.45);
\coordinate (j2) at (0.35,0.45);
\coordinate (j3) at (0.90,-0.20);
\coordinate (j4) at (-0.50,-0.85);
\coordinate (j5) at (0.25,-1.00);
\draw[dashed,gray!70] (-1,-1) rectangle (1,1);
\draw[{Stealth[length=1.6mm]}-{Stealth[length=1.6mm]},gray!70]
  (-1,1.14) -- (1,1.14) node[midway,above,gray,font=\scriptsize]
  {$2\delta_{\max}$};
\draw[gray!70] (j1) -- (j2);
\draw[gray!70] (j3) -- (j5);
\draw[gray!70] (j4) -- (j5);
\foreach \j in {j1,j3,j4,j5}
  \draw[-{Stealth[length=1.8mm]},gray!70] (i) -- (\j);
\draw[-{Stealth[length=2mm]},blue!70!black,thick] (i) -- (j2);
\foreach \j in {j1,j3,j4,j5} \node[vertex] at (\j) {};
\node[vertex,fill=blue!70!black] at (j2) {};
\node[blue!70!black] at ($(j2)+(0.15,0.11)$) {$j$};
\node[circle,fill=blue!70!black,inner sep=2.2pt] at (i) {};
\node[blue!70!black] at ($(i)+(0.16,-0.13)$) {$i$};
\node at (0,-1.28) {offsets $\bz_j-\bz_i$};
\end{scope}
\draw[-{Stealth[length=2.5mm]},thick] (1.85,0) -- (3.05,0)
   node[midway,above] {$\bu_{ij}$}
   node[midway,below,font=\scriptsize] {\eqref{eq:pseudo}};
\begin{scope}[shift={(3.5,-1.4)},scale=2.8]
  \fill[blue!12] (0.5,0.5) rectangle (0.75,0.75);
  \draw[step=0.25,gray!50,thin] (0,0) grid (1,1);
  \draw[thick] (0,0) rectangle (1,1);
  \coordinate (c) at (0.5,0.5);
  \foreach \p in {(0.05,0.725),(0.95,0.40),(0.25,0.075),(0.625,0.0)} {
     \draw[-{Stealth[length=1.4mm]},gray!70,shorten >=2pt] (c) -- \p;
     \node[vertex,fill=gray!70] at \p {};
  }
  \draw[-{Stealth[length=1.6mm]},blue!70!black,thick,shorten >=2pt]
    (c) -- (0.675,0.725);
  \node[vertex,fill=blue!70!black] (u) at (0.675,0.725) {};
  \node[blue!70!black] at ($(u)+(0.05,-0.055)$) {$j$};
  \node[blue!70!black,font=\scriptsize] at (0.55,0.84) {$\bu_{ij}$};
  \node[circle,fill=blue!70!black,inner sep=2.2pt] at (c) {};
  \node[blue!70!black] at ($(c)+(0.065,-0.055)$) {$i$};
  \node[font=\scriptsize,anchor=north east,black!60] at (0.475,0.465)
    {$\tfrac12\mathbf 1$};
  \foreach \p in {(0.5,0.5),(0.75,0.5),(0.5,0.75),(0.75,0.75)}
     \draw[blue!70!black,thick] \p circle (0.035);
  \node at (0.5,-0.125) {knot grid on $[0,1]^2$, $k{=}5$};
\end{scope}
\end{tikzpicture}}
\end{minipage}\hfill
\begin{minipage}[c]{0.47\linewidth}
\caption{Construction of two dimensional pseudo-coordinates for $D{=}2$.
\textbf{Left:} The convolution at node $i$ represents each neighbour $j$
by its spatial offset $\bz_j-\bz_i$. The dashed box marks the largest offset
magnitude $\delta_{\max}$ in the graph. \textbf{Right:} The map in
\eqref{eq:pseudo} rescales this box to the unit square. Node $i$ lands at the
centre $\tfrac12\mathbf 1$, node $j$ lands at $\bu_{ij}$ in blue, and the
largest offset reaches the boundary.
}
\label{fig:coords}
\end{minipage}
\end{figure}

\paragraph{Graphs and pseudo-coordinates.}
We consider graphs $G=(V,E)$ embedded in a coordinate space: every node $i$
carries features $\bx_i\in\R^{\Cin}$ and a spatial position $\bz_i\in\R^D$
(2-D image keypoints, vertices of a 3-D mesh, \dots). The convolution below
depends on \emph{relative} positions.
Because the kernel will be parameterised by basis functions living on a \emph{fixed, bounded}
domain, the offsets are affinely rescaled into the unit cube: with
$\delta_{\max}$ the largest absolute offset component occurring in the
graph, the \emph{pseudo-coordinate} of edge $(j,i)$ is
\begin{equation}
\bu_{ij} \;=\; \frac{\bz_j-\bz_i}{2\,\delta_{\max}} \;+\; \tfrac12\mathbf 1
\;\;\in\;[0,1]^D
\label{eq:pseudo}
\end{equation}
The zero offset lands at the centre $\tfrac12\mathbf 1$, a neighbour to the left/below has
$u_d \leq \tfrac12$ in that dimension, and the largest offset touches the
boundary.
See Figure~\ref{fig:coords} for an illustration of the pseudo-coordinate construction.
These pseudo-coordinates are the only way geometry enters the model.
Polar or spherical encodings can be handled analogously.

\emph{Angular and spherical coordinates:} Rescaling periodic coordinates into the unit cube would introduce a seam at $0\equiv2\pi$ and, for spherical angles, singularities at the poles.
Angular coordinates can be embedded as $(\cos\theta,\sin\theta)$ and directions as unit vectors $\mathbf n\in S^2$.

\paragraph{Continuous convolution kernels.}
An image convolution assigns a separate weight matrix to each of the
finitely many pixel offsets. On an irregular graph the offsets $\bu_{ij}$
vary continuously, so we instead learn a \emph{kernel function}
$g:[0,1]^D\to\R^{\Cin\times\Cout}$ and evaluate it at the observed
pseudo-coordinates. The kernel is parameterised as a linear combination
$g(\bu)=\sum_{p=1}^{K}B_p(\bu)\,\Theta_p$ of $K$ scalar basis
functions $B_p\colon[0,1]^D\to\R$ with learnable weight matrices
$\Theta_p\in\R^{\Cin\times\Cout}$, and a layer computes
\begin{equation}
\bx'_i \;=\; \Theta_{\mathrm{root}}\,\bx_i \;+\;
\mathop{\square}_{j\in\mathcal N(i)}\;\Bigl(\sum_{p=1}^{K} B_p(\bu_{ij})\,\Theta_p\Bigr)\bx_j \;+\; \mathbf b,
\label{eq:splineconv}
\end{equation}
with a separate root weight for the node itself, a bias, and aggregation
$\square \in \{\mathrm{mean}, \mathrm{add}, \mathrm{max}\}$ over the neighbourhood.
This operator was introduced by SplineCNN~\citep{fey2018splinecnn}.
The difference lies in the choice of the basis functions $B_p$, which are fixed B-splines there (see the remark below) and learnable rational functions for \ours{}.

\setcounter{topnumber}{1}

\begin{figure}[t]
\begin{minipage}[c]{0.52\linewidth}
\centering
\begin{tikzpicture}[
  font=\scriptsize,
  declare function={
    numP(\t)=1-\t*\t;
    denQ(\t)=1.5*\t*\t+\t-0.8;
  },
]
\begin{axis}[
  name=polys,
  width=0.455\linewidth, height=4.4cm,
  xmin=-1, xmax=1, ymin=-1.4, ymax=2.1,
  xtick={-1,0,1}, ytick={-1,0,1,2},
  xlabel={$t$},
  every axis plot/.append style={thick},
  tick label style={font=\tiny}, label style={font=\scriptsize},
  xlabel style={yshift=1.5mm},
  title={\scriptsize (a) polynomials $P$, $Q$},
]
\addplot[gray!60, domain=-1:1, samples=100] {denQ(x)};
\addplot[blue!70!black, domain=-1:1, samples=100] {numP(x)};
\fill[black] (axis cs:0.4694,0) circle (1.4pt);
\node[font=\tiny, anchor=north west, inner sep=1pt] at (axis cs:0.5,-0.05)
  {$t^\ast$};
\node[font=\tiny, blue!70!black] at (axis cs:-0.42,1.15) {$P$};
\node[font=\tiny, gray!45!black] at (axis cs:-0.72,-0.85) {$Q$};
\end{axis}
\begin{axis}[
  at={(polys.outer east)}, anchor=outer west, xshift=-1mm,
  name=plainrat,
  width=0.455\linewidth, height=4.4cm,
  xmin=-1, xmax=1, ymin=-4, ymax=4,
  xtick={-1,0,1}, ytick={-4,-2,0,2,4},
  xlabel={$t$},
  every axis plot/.append style={thick},
  tick label style={font=\tiny}, label style={font=\scriptsize},
  xlabel style={yshift=1.5mm},
  title={\scriptsize (b) rational $R=P/Q$},
]
\draw[dashed,black!60] (axis cs:0.4694,-4) -- (axis cs:0.4694,4);
\addplot[blue!70!black, domain=-1:0.45, samples=120,
  restrict y to domain=-4:4] {numP(x)/denQ(x)};
\addplot[blue!70!black, domain=0.49:1, samples=120,
  restrict y to domain=-4:4] {numP(x)/denQ(x)};
\node[font=\tiny, anchor=east] at (axis cs:0.42,-3.2)
  {pole at $t^\ast$};
\end{axis}
\begin{axis}[
  at={(plainrat.outer east)}, anchor=outer west, xshift=-1mm,
  width=0.455\linewidth, height=4.4cm,
  xmin=-1, xmax=1, ymin=0, ymax=3,
  xtick={-1,0,1}, ytick={0,1,2,3},
  xlabel={$t$},
  every axis plot/.append style={thick},
  tick label style={font=\tiny}, label style={font=\scriptsize},
  xlabel style={yshift=1.5mm},
  title={\scriptsize (c) safe Pad\'e $B$},
]
\draw[dashed,black!60] (axis cs:-1,1) -- (axis cs:1,1);
\addplot[gray!60, domain=-1:1, samples=200] {1+abs(denQ(x))};
\addplot[blue!70!black, domain=-1:1, samples=200]
  {numP(x)/(1+abs(denQ(x)))};
\node[font=\tiny, gray!45!black, anchor=north west, inner sep=1pt] at (axis cs:-0.95,2.9)
  {$1{+}|Q|\ge 1$};
\node[font=\tiny, blue!70!black] at (axis cs:0.05,0.82) {$B$};
\end{axis}
\end{tikzpicture}
\end{minipage}\hfill
\begin{minipage}[c]{0.45\linewidth}
\caption{From polynomials to safe Pad\'e basis functions,
Eqs.~\eqref{eq:polynomial}--\eqref{eq:safeform}, with the same numerator
$P$ and denominator $Q$ in all three panels. \textbf{(a)}~Two polynomials
with learnable coefficients; $Q$ has a root $t^\ast$ inside the domain.
\textbf{(b)}~The plain rational function $R=P/Q$ diverges at $t^\ast$,
and any root of $Q$ can move into the domain during training.
\textbf{(c)}~In the safe form the denominator $1+|Q|$ (gray) never drops
below one (dashed), so $B$ stays bounded and well defined for every
setting of the coefficients; where $Q$ vanishes, $B$ simply equals $P$.}
\label{fig:pade}
\end{minipage}
\end{figure}
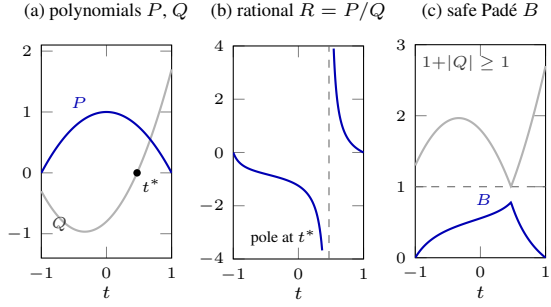

\paragraph{Rational safe Pad\'e basis functions.}
In \ours{} the scalar basis functions $B_p\colon[0,1]^D\to\R$ of \eqref{eq:splineconv} are learnable rational functions.
Each $B_p$ is a quotient of two polynomials whose coefficients are network parameters.
We build up the functional form in one dimension first.
A \emph{polynomial} of degree $m$ with learnable coefficients $a_0,\dots,a_m$ is
\begin{equation}
P(t) \;=\; \sum_{j=0}^{m} a_j\,t^j.
\label{eq:polynomial}
\end{equation}
A \emph{rational} function of degrees $(m,n)$ is a quotient of two such
polynomials,
\begin{equation}
R(t) \;=\; \frac{P(t)}{Q(t)}
\;=\; \frac{\sum_{j=0}^{m} a_j\,t^j}{\sum_{j=0}^{n} b_j\,t^j}.
\label{eq:rational}
\end{equation}
Training a free denominator is unstable, since it can become zero, making
the evaluation undefined, but also being close to zero will lead to numerical overflow.
We therefore use the \emph{safe} Pad\'e form~\citep{molina2020pade}, in which the denominator polynomial $Q$ (without constant term) enters only through $1+|Q(\cdot)|\ge 1$:
\begin{equation}
B(t) \;=\; \frac{P(t)}{1+\bigl|Q(t)\bigr|}
\;=\; \frac{\sum_{j=0}^{m} a_j\,t^j}
           {1+\Bigl|\sum_{j=1}^{n} b_j\,t^j\Bigr|},
\label{eq:safeform}
\end{equation}
which remains well defined and bounded for every value of the learnt coefficients $a_j,b_j$.
Figure~\ref{fig:pade} contrasts the plain rational~\eqref{eq:rational} with the safe form~\eqref{eq:safeform}.

In practice we expand $P$ and $Q$ not in the monomial basis $1,t,t^2,\dots$ but in Chebyshev polynomials of the first kind $\phi_0,\phi_1,\dots$ ($\phi_0{=}1$, $\phi_1{=}t$, $\phi_{j+1}{=}2t\phi_j-\phi_{j-1}$), i.e.\ $P(t)=\sum_j a_j\,\phi_j(t)$.
This parametrization spans the same function class but is better conditioned on bounded domains.
The map from monomial coefficients to polynomial values on an interval is exponentially ill-conditioned in the degree~\citep{gautschi1979condition}, so with monomials a small step in a high-order coefficient barely moves the function near $t=0$ but changes it violently near the boundary, giving gradients on wildly different scales.
The Chebyshev polynomials, by contrast, are orthogonal and uniformly bounded, $|\phi_j(t)|\le 1$ on $[-1,1]$, so every coefficient acts on a comparable scale and $|a_j|$ directly bounds the contribution of the $j$-th term.

Below we extend this one-dimensional basis function to multiple dimensions in two different ways.

\paragraph{Product rational basis.}
Each of the $K$ basis functions factorises over the coordinate dimensions into univariate safe Pad\'e functions
\eqref{eq:safeform},
\begin{equation}
B_{\mathbf p}(\bu) \;=\; \prod_{d=1}^{D}
\frac{P_{d,p}(u_d)}{1+\bigl|Q_{d,p}(u_d)\bigr|},
\label{eq:safepade}
\end{equation}
where every factor has its own learnable polynomials $P_{d,p}$ and $Q_{d,p}$.

\paragraph{Multivariate rational basis.}
Each basis function $B_p\colon[0,1]^D\to\R$ is a single non-separable rational function of all coordinates jointly:
\begin{equation}
B_p(\bu) = \frac{P_p(\bu)}{1+|Q_p(\bu)|},\quad
P_p(\bu)=\!\!\sum_{|\alpha|\le m}\! a_{p,\alpha}\,
u_1^{\alpha_1}\!\cdots u_D^{\alpha_D},\quad
Q_p(\bu)=\!\!\sum_{1\le|\alpha|\le n}\! b_{p,\alpha}\,
u_1^{\alpha_1}\!\cdots u_D^{\alpha_D},
\label{eq:multivariate}
\end{equation}
with multi-indices $\alpha\in\mathbb N_0^D$ of total degree $|\alpha|\le m$ and $|\alpha|\le n$. As in the univariate case, the implementation uses the Chebyshev parametrization. Each monomial $u_1^{\alpha_1}\!\cdots u_D^{\alpha_D}$ is replaced by the product $\phi_{\alpha_1}(t_1)\cdots\phi_{\alpha_D}(t_D)$, spanning the same function class in better-conditioned form.

\paragraph{Original SplineCNN and B-spline basis.}
The original SplineCNN~\citep{fey2018splinecnn} uses the operator~\eqref{eq:splineconv}, but chooses the basis functions $B_p$ as B-splines, see Figure~\ref{fig:teaser}. This construction yields spatial localisation and a strong geometric inductive bias. However, in every coordinate at most 2 B-splines are active, which means in larger dimensions that all but $2^D$ of the $\Theta_{\mathbf p}$ receive zero gradient. This results in sparse, uneven updates and dense evaluation wasted on zeros. The basis size $K=k^D$ grows exponentially with the coordinate dimension $D$.

\begin{table}[!b]
\centering\footnotesize
\vspace{5pt}
\caption{Semantic keypoint matching on PascalVOC and SPair-71k. Accuracy (\%) follows
the protocol of each pipeline and is the mean $\pm$ std over 5 seeds, and
\ranklegend{} mark the three highest means per column. Parameter counts exclude
the VGG16 backbone and cover the whole DGMC model but only the NMT
graph-convolution layers, which are larger because they use 648 channels
instead of the 256 and 128 of DGMC. All other columns report the best test
epoch.}
\label{tab:keypoint}
\setlength{\tabcolsep}{4pt}
\renewcommand{\arraystretch}{1.00}
\resizebox{\ifdim\width>\textwidth\textwidth\else\width\fi}{!}{%
\begin{tabular}{lrccrcc}
\toprule
 & \multicolumn{3}{c}{DGMC~\citep{fey2020dgmc}} & \multicolumn{3}{c}{NMT~\citep{normmatchtrans}} \\
\cmidrule(lr){2-4}\cmidrule(lr){5-7}
Method & Params $\downarrow$ & PascalVOC $\uparrow$ & SPair-71k $\uparrow$
       & Params $\downarrow$ & PascalVOC $\uparrow$ & SPair-71k $\uparrow$ \\
\midrule
\rowcolor[HTML]{EDEDED}
\multicolumn{7}{l}{\textit{Spline baselines}} \\
SplineCNN ($K{=}25$) & 9.50M & $73.00 \pm 0.40$ & $76.18 \pm 0.64$ & 28.17M & \yellowc $83.10 \pm 0.36$ & \orangec $82.10 \pm 0.11$ \\
SplineCNN ($K{=}9$) & 3.74M & $72.78 \pm 0.40$ & $76.04 \pm 0.43$ & 10.84M & $82.66 \pm 0.30$ & $81.54 \pm 0.31$ \\
SplineCNN ($K{=}4$) & 1.93M & $71.48 \pm 0.40$ & $71.92 \pm 0.38$ &  5.42M & $80.10 \pm 0.46$ & $77.79 \pm 0.40$ \\
\rowcolor[HTML]{EDEDED}
\multicolumn{7}{l}{\textit{MLP learned-basis control}} \\
MLP basis ($K{=}4$) & 1.94M & $72.76 \pm 0.27$ & $73.74 \pm 0.71$ &  5.42M & $81.90 \pm 0.27$ & $80.40 \pm 0.57$ \\
MLP basis ($K{=}9$) & 3.74M & $72.04 \pm 0.73$ & $74.34 \pm 0.62$ & 10.84M & $82.12 \pm 0.51$ & $81.05 \pm 0.32$ \\
\rowcolor[HTML]{EDEDED}
\multicolumn{7}{l}{\textit{RBF-GNN with a product basis}} \\
\textbf{RBF-GNN} ($K{=}25$) & 9.50M & $72.88 \pm 0.48$ & $76.34 \pm 0.42$ & 28.17M & \redc $83.78 \pm 0.35$ & \yellowc $82.08 \pm 0.21$ \\
\textbf{RBF-GNN} ($K{=}9$) & 3.74M & $73.26 \pm 0.18$ & $76.24 \pm 0.73$ & 10.84M & $82.85 \pm 0.42$ & $81.64 \pm 0.35$ \\
\rowcolor[HTML]{EDEDED}
\multicolumn{7}{l}{\textit{RBF-GNN with a multivariate basis}} \\
\textbf{RBF-GNN} ($K{=}9$), spline init & 3.74M & \redc $74.12 \pm 0.48$ & \orangec $77.68 \pm 0.35$ & 10.84M & $83.08 \pm 0.26$ & \redc $82.11 \pm 0.35$ \\
\textbf{RBF-GNN} ($K{=}4$), PCA$+$vp & 1.94M & \yellowc $73.78 \pm 0.41$ & $77.44 \pm 0.36$ &  5.42M & $82.42 \pm 0.43$ & $81.52 \pm 0.22$ \\
\textbf{RBF-GNN} ($K{=}6$), PCA$+$vp & 2.66M & \orangec $74.06 \pm 0.37$ & \redc $77.80 \pm 0.66$ &  7.59M & $82.95 \pm 0.42$ & $81.71 \pm 0.18$ \\
\textbf{RBF-GNN} ($K{=}9$), PCA$+$vp & 3.74M & $73.72 \pm 0.19$ & \yellowc $77.50 \pm 0.27$ & 10.84M & $82.88 \pm 0.38$ & $81.59 \pm 0.14$ \\
\rowcolor[HTML]{EDEDED}
\multicolumn{7}{l}{\textit{RBF-GNN with a multivariate basis, degree control}} \\
\textbf{RBF-GNN} ($K{=}25$), degrees $(5,4)$ & 9.51M & $73.08 \pm 0.25$ & $76.26 \pm 0.23$ & 28.17M & \orangec $83.35 \pm 0.38$ & $82.01 \pm 0.21$ \\
\bottomrule
\end{tabular}}
\end{table}

\begin{figure}[t]
\centering
\resizebox{0.96\linewidth}{!}{%
\begin{tikzpicture}[
  kp/.style={circle, inner sep=0pt, minimum size=2.6pt, draw=white, line width=0.3pt},
  match/.style={line width=0.7pt, opacity=0.75},
  corr/.style={densely dashed, black!70, line width=0.5pt},
  mdot/.style={circle, inner sep=0pt, minimum size=2.8pt, draw=black, fill=white, line width=0.4pt}]

\node[anchor=south west, inner sep=0] (c1) at (0,0)
  {\includegraphics[height=22mm]{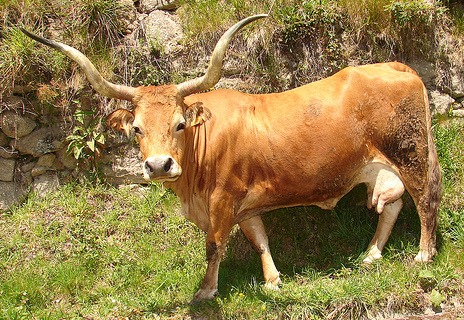}};
\node[anchor=south west, inner sep=0] (c2) at ($(c1.south east)+(3mm,0)$)
  {\includegraphics[height=28mm]{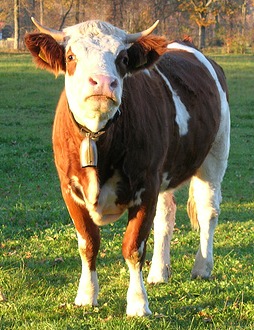}};
\begin{scope}[shift={(c1.south west)}, x=31.90mm, y=22mm]
  \node[kp, fill=red]              (a-hornL) at (0.052,0.903) {};
  \node[kp, fill=orange]           (a-hornR) at (0.560,0.947) {};
  \node[kp, fill=yellow!90!black]  (a-eyeL)  at (0.291,0.594) {};
  \node[kp, fill=green!65!black]   (a-eyeR)  at (0.386,0.609) {};
  \node[kp, fill=cyan!70!black]    (a-nose)  at (0.343,0.444) {};
  \node[kp, fill=blue!70]          (a-back)  at (0.847,0.797) {};
  \node[kp, fill=violet]           (a-hoofF) at (0.440,0.100) {};
  \node[kp, fill=magenta!80!black] (a-hoofB) at (0.909,0.203) {};
\end{scope}
\begin{scope}[shift={(c2.south west)}, x=21.55mm, y=28mm]
  \node[kp, fill=red]              (b-hornL) at (0.134,0.936) {};
  \node[kp, fill=orange]           (b-hornR) at (0.610,0.924) {};
  \node[kp, fill=yellow!90!black]  (b-eyeL)  at (0.272,0.824) {};
  \node[kp, fill=green!65!black]   (b-eyeR)  at (0.484,0.818) {};
  \node[kp, fill=cyan!70!black]    (b-nose)  at (0.402,0.709) {};
  \node[kp, fill=blue!70]          (b-back)  at (0.756,0.852) {};
  \node[kp, fill=violet]           (b-hoofF) at (0.543,0.067) {};
  \node[kp, fill=magenta!80!black] (b-hoofB) at (0.787,0.191) {};
\end{scope}
\draw[match, red]              (a-hornL) -- (b-hornL);
\draw[match, orange]           (a-hornR) -- (b-hornR);
\draw[match, yellow!90!black]  (a-eyeL)  -- (b-eyeL);
\draw[match, green!65!black]   (a-eyeR)  -- (b-eyeR);
\draw[match, cyan!70!black]    (a-nose)  -- (b-nose);
\draw[match, blue!70]          (a-back)  -- (b-back);
\draw[match, violet]           (a-hoofF) -- (b-hoofF);
\draw[match, magenta!80!black] (a-hoofB) -- (b-hoofB);
\coordinate (capa) at ($(c1.south west)!0.5!(c2.south east)$);
\node[font=\small, anchor=north] at (capa |- 0,-6.5mm)
  {(a) semantic keypoint matching};

\node[anchor=south west, inner sep=0] (p1) at ($(c2.south east)+(9mm,0)$)
  {\includegraphics[height=28mm]{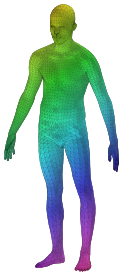}};
\node[anchor=south west, inner sep=0] (p2) at ($(p1.south east)+(10mm,0)$)
  {\includegraphics[height=28mm]{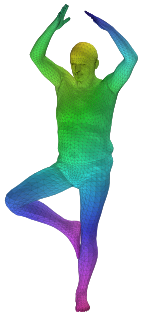}};
\begin{scope}[shift={(p1.south west)}, x=12.48mm, y=28mm]
  \coordinate (fa-head)  at (0.5058,0.9845);
  \coordinate (fa-handR) at (0.9598,0.4087);
  \coordinate (fa-foot)  at (0.2949,0.0329);
\end{scope}
\begin{scope}[shift={(p2.south west)}, x=12.58mm, y=28mm]
  \coordinate (fb-head)  at (0.5819,0.8669);
  \coordinate (fb-handR) at (0.6497,0.9540);
  \coordinate (fb-foot)  at (0.5657,0.2107);
\end{scope}
\draw[corr] (fa-head)  to[bend left=30]  (fb-head);
\draw[corr] (fa-handR) to[bend left=20]  (fb-handR);
\draw[corr] (fa-foot)  to[bend right=25] (fb-foot);
\foreach \pt in {fa-head,fa-handR,fa-foot,fb-head,fb-handR,fb-foot}
  \node[mdot] at (\pt) {};
\coordinate (capb) at ($(p1.south west)!0.5!(p2.south east)$);
\node[font=\small, anchor=north] at (capb |- 0,-6.5mm)
  {(b) shape correspondence};

\begin{axis}[
  at={($(p2.south east)+(3mm,-5mm)$)}, anchor=south west,
  width=74mm, height=52mm,
  view={-78}{16},
  hide axis, clip=false,
  unit vector ratio=1.8 1 1,
  xmin=0, xmax=1, ymin=0, ymax=1, zmin=0, zmax=0.55]
  \coordinate (fo) at (axis cs:0,1,0);
  \coordinate (fx) at (axis cs:0,0,0);
  \coordinate (fz) at (axis cs:0,1,0.55);
  \addplot3[black!60, line width=0.4pt] coordinates
    {(0,0,0) (0,1,0) (0,1,0.55) (0,0,0.55) (0,0,0)};
  \addplot3[black!30, densely dotted] coordinates
    {(1,0,0) (1,1,0) (1,1,0.55) (1,0,0.55) (1,0,0)};
  \addplot3[black!30, densely dotted] coordinates {(0,0,0) (1,0,0)};
  \addplot3[black!30, densely dotted] coordinates {(0,1,0) (1,1,0)};
  \addplot3[black!30, densely dotted] coordinates {(0,0,0.55) (1,0,0.55)};
  \addplot3[black!30, densely dotted] coordinates {(0,1,0.55) (1,1,0.55)};
  \addplot3[only marks, mark=*, mark size=0.3pt, color=red!80!black, opacity=0.35]
    table[x=t, y expr=1-\thisrow{x}, z=y] {figures/events_on.dat};
  \addplot3[only marks, mark=*, mark size=0.3pt, color=blue!75!black, opacity=0.35]
    table[x=t, y expr=1-\thisrow{x}, z=y] {figures/events_off.dat};
  \addplot3[fill=black!8, fill opacity=0.55, draw=black!45, line width=0.35pt] coordinates
    {(0.72,0,0) (0.72,1,0) (0.72,1,0.55) (0.72,0,0.55) (0.72,0,0)};
  \addplot3[only marks, mark=*, mark size=0.15pt, color=red!85!black]
    table[x=t, y expr=1-\thisrow{x}, z=y] {figures/events_slice_on.dat};
  \addplot3[only marks, mark=*, mark size=0.15pt, color=blue!80!black]
    table[x=t, y expr=1-\thisrow{x}, z=y] {figures/events_slice_off.dat};
  \addplot3[mark=none, black, line width=0.55pt, empty line=jump]
    table[x=t, y expr=1-\thisrow{x}, z=y] {figures/eventedges.dat};
  \addplot3[only marks, mark=*, mark size=1.4pt, color=orange!90!black]
    coordinates {(0.72,0.4507,0.2718)};
  \draw[-{Stealth[length=1.5mm]}, black!70] (axis cs:0,0,0) -- (axis cs:1.12,0,0)
    node[pos=1, below right, inner sep=1pt, font=\scriptsize] {$t$};
  \draw[-{Stealth[length=1.3mm]}, black!70, line width=0.5pt]
    (axis cs:0,0.30,-0.03) -- (axis cs:0,0.62,-0.03);
  \coordinate (axsw) at (axis cs:0,1,0);
  \coordinate (axse) at (axis cs:1,0,0);
\end{axis}
\path let \p1=(fo), \p2=($(fx)-(fo)$), \p3=($(fz)-(fo)$) in
  node[inner sep=0, anchor=south west, transform shape,
       cm={\x2/1cm,\y2/1cm,\x3/1cm,\y3/1cm,(\p1)}] at (0,0)
  {\includegraphics[width=1cm,height=1cm]{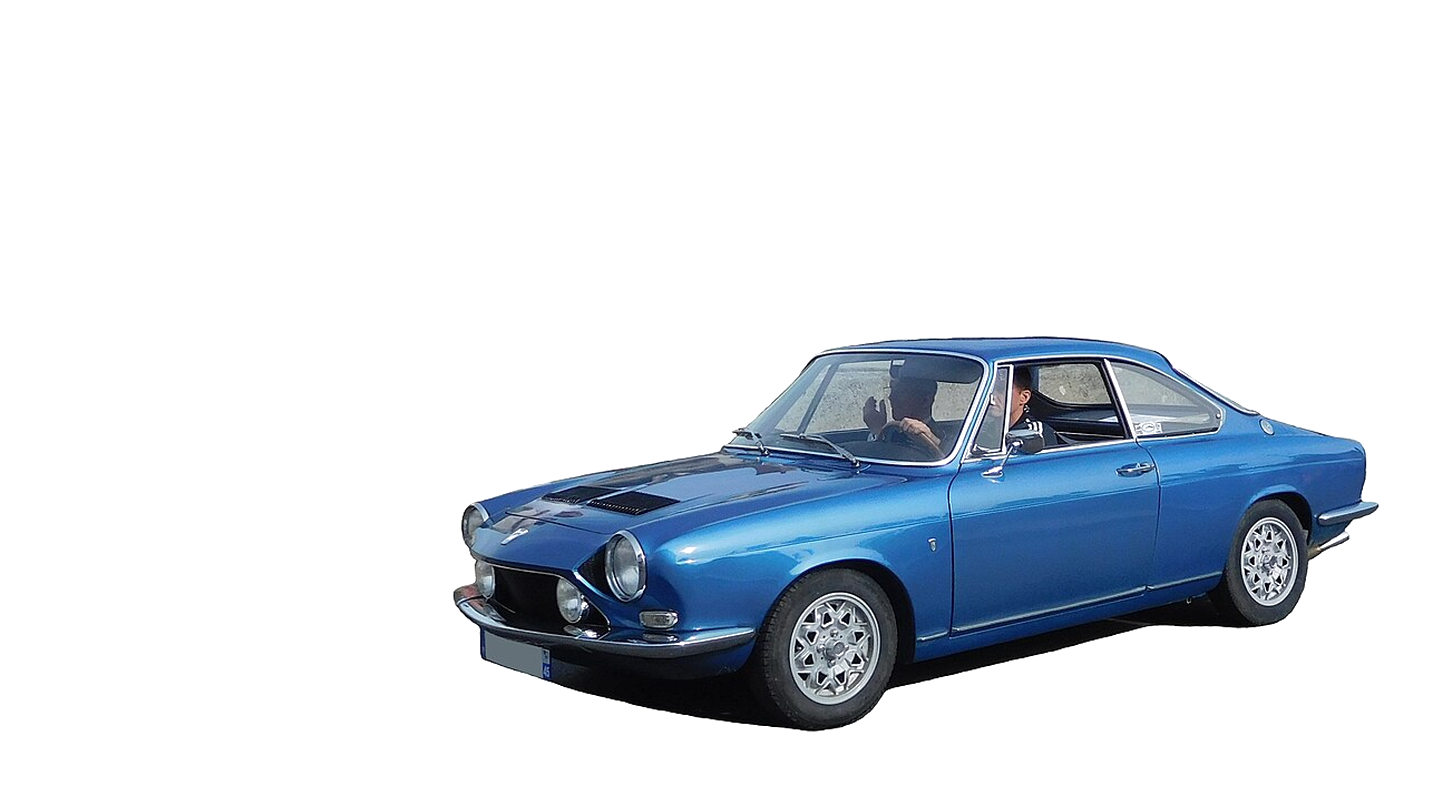}};
\coordinate (capc) at ($(axsw)!0.5!(axse)$);
\node[font=\small, anchor=north] at (capc |- 0,-6.5mm)
  {(c) event-based vision};
\end{tikzpicture}}
\caption{Tasks with pseudo-coordinate graphs used in our experiments. \textbf{(a)}~Semantic keypoint matching on SPair-71k: keypoints of two images of the same category are graph nodes, and corresponding keypoints (same color, 8 of 20 shown) are matched across instances with different viewpoint and appearance. \textbf{(b)}~Shape correspondence on FAUST: every vertex of a triangulated human scan is matched to its vertex on another person in a different pose; colors encode the ground-truth correspondence, and dashed arcs show example matches. \textbf{(c)}~Event-based vision: object moving through a scene triggers a stream of events. The events of a short time window (gray plane) trace the contour of the car; each event is a graph node (orange) connected to its space-time neighbors, and the pseudo-coordinates are relative $(x,y,t)$ offsets.
}
\label{fig:tasks}
\end{figure}
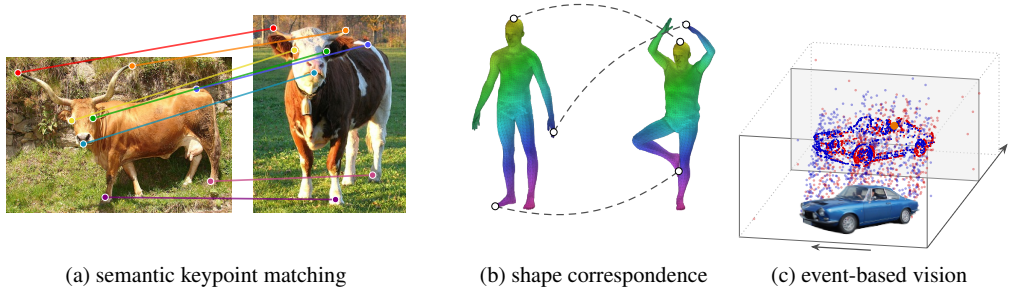

\begin{table}[t]
\centering
\begin{minipage}[t][4\baselineskip][t]{0.49\textwidth}
\caption{
Accuracy (\%) for shape correspondence on FAUST.
We could not fully reproduce the published numbers.
}
\label{tab:faust}
\end{minipage}\hfill
\begin{minipage}[t][4\baselineskip][t]{0.49\textwidth}
\caption{Instance-average mIoU (\%) for part segmentation on ShapeNet-Part with 6-D
position and normal pseudo-coordinates.}
\label{tab:shapenet-part}
\end{minipage}\par\nointerlineskip
\begin{minipage}[t]{0.49\textwidth}
\vspace{0pt}
\centering\fontsize{7.4pt}{8.2pt}\selectfont
\setlength{\tabcolsep}{4pt}
\renewcommand{\arraystretch}{1.00}
\begin{tabular*}{\linewidth}{l@{\extracolsep{\fill}}rc}
\toprule
Method & Params $\downarrow$ & Final $\uparrow$ \\
\midrule
\rowcolor[HTML]{EDEDED}
\multicolumn{3}{l}{\textit{Spline baselines}} \\
\emph{SplineCNN ($K{=}125$), published} & 4.10M & \emph{99.20} \\
SplineCNN ($K{=}125$), mean $+$ clip & 4.11M & $98.71 \pm 0.08$ \\
SplineCNN ($K{=}125$), add $+$ clip & 4.11M & $97.23 \pm 0.51$ \\
SplineCNN ($K{=}125$), add, no clip & 4.11M & $29.1 \pm 50.3$ \\
SplineCNN ($K{=}27$) & 2.30M & $83.6 \pm 15.6$ \\
SplineCNN ($K{=}8$) & \orangec 1.95M & $3.4 \pm 3.3$ \\
\rowcolor[HTML]{EDEDED}
\multicolumn{3}{l}{\textit{MLP learned-basis control}} \\
MLP basis ($K{=}8$) & \yellowc 1.96M & $98.21 \pm 1.11$ \\
\rowcolor[HTML]{EDEDED}
\multicolumn{3}{l}{\textit{RBF-GNN with a product basis}} \\
\textbf{RBF-GNN} ($K{=}27$) & 2.30M & \yellowc $99.24 \pm 0.08$ \\
\rowcolor[HTML]{EDEDED}
\multicolumn{3}{l}{\textit{RBF-GNN with a multivariate basis}} \\
\textbf{RBF-GNN} ($K{=}4$) & \redc 1.89M & \orangec $99.38 \pm 0.09$ \\
\textbf{RBF-GNN} ($K{=}4$), mean aggr. & \redc 1.89M & $98.93 \pm 0.10$ \\
\textbf{RBF-GNN} ($K{=}8$) & 1.97M & \redc $99.41 \pm 0.04$ \\
\bottomrule
\end{tabular*}
\end{minipage}\hfill
\begin{minipage}[t]{0.49\textwidth}
\vspace{0pt}
\centering\fontsize{7.4pt}{8.2pt}\selectfont
\setlength{\tabcolsep}{5pt}
\renewcommand{\arraystretch}{1.00}
\begin{tabular*}{\linewidth}{l@{\extracolsep{\fill}}rc}
\toprule
Method & Params $\downarrow$ & Final mIoU $\uparrow$ \\
\midrule
\rowcolor[HTML]{EDEDED}
\multicolumn{3}{l}{\textit{Spline baselines}} \\
SplineCNN ($K{=}729$) & 1.60M & \redc $65.12 \pm 0.41$ \\
SplineCNN ($K{=}64$) & 177.91k & $60.02 \pm 0.23$ \\
\rowcolor[HTML]{EDEDED}
\multicolumn{3}{l}{\textit{MLP learned-basis control}} \\
MLP basis ($K{=}8$) & \redc 60.75k & $62.80 \pm 0.45$ \\
MLP basis ($K{=}32$) & \yellowc 116.88k & $63.81 \pm 0.46$ \\
MLP basis ($K{=}64$) & 191.73k & $64.41 \pm 0.40$ \\
MLP basis ($K{=}128$) & 341.43k & $63.78 \pm 0.17$ \\
MLP basis ($K{=}256$) & 640.82k & $64.05 \pm 0.75$ \\
\rowcolor[HTML]{EDEDED}
\multicolumn{3}{l}{\textit{RBF-GNN with a multivariate basis}} \\
\textbf{RBF-GNN} ($K{=}8$) & \orangec 64.87k & $61.88 \pm 0.38$ \\
\textbf{RBF-GNN} ($K{=}32$) & 137.43k & \yellowc $64.86 \pm 0.18$ \\
\textbf{RBF-GNN} ($K{=}64$) & 234.16k & $64.68 \pm 0.55$ \\
\textbf{RBF-GNN} ($K{=}128$) & 427.63k & \orangec $64.94 \pm 0.70$ \\
\textbf{RBF-GNN} ($K{=}256$) & 814.58k & $64.69 \pm 0.41$ \\
\bottomrule
\end{tabular*}
\end{minipage}
\end{table}

\begin{table}[t]
\centering
\begin{minipage}[t][4\baselineskip][t]{0.49\textwidth}
\caption{Test accuracy (\%) for object recognition on N-Caltech101 with AEGNN.}
\label{tab:ncaltech101}
\end{minipage}\hfill
\begin{minipage}[t][4\baselineskip][t]{0.49\textwidth}
\caption{Test accuracy (\%) for car recognition on N-Cars with AEGNN.}
\label{tab:ncars}
\end{minipage}\par\nointerlineskip
\begin{minipage}[t]{0.49\textwidth}
\vspace{0pt}
\centering\fontsize{8.2pt}{9.1pt}\selectfont
\setlength{\tabcolsep}{5pt}
\renewcommand{\arraystretch}{1.00}
\begin{tabular*}{\linewidth}{l@{\extracolsep{\fill}}crc}
\toprule
Method & Aggr. & Params $\downarrow$ & Final $\uparrow$ \\
\midrule
\rowcolor[HTML]{EDEDED}
\multicolumn{4}{l}{\textit{Spline baselines}} \\
SplineCNN ($K{=}8$) & mean & \orangec 77.68k & $47.35 \pm 0.15$ \\
SplineCNN ($K{=}8$) & max & \orangec 77.68k & \orangec $56.00 \pm 0.92$ \\
\rowcolor[HTML]{EDEDED}
\multicolumn{4}{l}{\textit{PointNet learned-kernel control}} \\
PointNet & max & \redc 55.68k & $54.99 \pm 0.84$ \\
\rowcolor[HTML]{EDEDED}
\multicolumn{4}{l}{\textit{RBF-GNN with a multivariate basis}} \\
\textbf{RBF-GNN} ($K{=}8$) & mean & \yellowc 91.57k & \yellowc $55.05 \pm 0.58$ \\
\textbf{RBF-GNN} ($K{=}8$) & max & \yellowc 91.57k & \redc $57.55 \pm 0.42$ \\
\bottomrule
\end{tabular*}
\end{minipage}\hfill
\begin{minipage}[t]{0.49\textwidth}
\vspace{0pt}
\centering\fontsize{8.2pt}{9.1pt}\selectfont
\setlength{\tabcolsep}{5pt}
\renewcommand{\arraystretch}{1.00}
\begin{tabular*}{\linewidth}{l@{\extracolsep{\fill}}crc}
\toprule
Method & Aggr. & Params $\downarrow$ & Final $\uparrow$ \\
\midrule
\rowcolor[HTML]{EDEDED}
\multicolumn{4}{l}{\textit{Spline baselines}} \\
SplineCNN ($K{=}8$) & mean & \yellowc 26.99k & $87.83 \pm 0.25$ \\
SplineCNN ($K{=}8$) & max & \yellowc 26.99k & $89.81 \pm 0.27$ \\
\rowcolor[HTML]{EDEDED}
\multicolumn{4}{l}{\textit{PointNet learned-kernel control}} \\
PointNet & max & \redc 4.99k & $87.17 \pm 0.24$ \\
\rowcolor[HTML]{EDEDED}
\multicolumn{4}{l}{\textit{RBF-GNN with a multivariate basis}} \\
\textbf{RBF-GNN} ($K{=}4$) & mean & \orangec 21.10k & \yellowc $90.24 \pm 0.39$ \\
\textbf{RBF-GNN} ($K{=}8$) & mean & 40.88k & \orangec $90.49 \pm 0.05$ \\
\textbf{RBF-GNN} ($K{=}8$) & max & 40.88k & \redc $91.00 \pm 0.43$ \\
\bottomrule
\end{tabular*}
\end{minipage}
\end{table}

\begin{table}[t]
\centering
\begin{minipage}[b]{0.49\textwidth}
\caption{Test accuracy (\%) for MNIST superpixel classification.}
\label{tab:mnist-superpixels}
\end{minipage}\hfill
\begin{minipage}[b]{0.49\textwidth}
\caption{Test accuracy (\%) with 5-D bilateral pseudo-coordinates on CIFAR-10
superpixels.}
\label{tab:cifar10-bilateral}
\end{minipage}\par\nointerlineskip
\begin{minipage}[t]{0.49\textwidth}
\vspace{0pt}
\centering\fontsize{7.4pt}{8.2pt}\selectfont
\setlength{\tabcolsep}{3.5pt}
\renewcommand{\arraystretch}{1.00}
\begin{tabular*}{\linewidth}{l@{\extracolsep{\fill}}rc}
\toprule
Method & Params $\downarrow$ & Final $\uparrow$ \\
\midrule
\rowcolor[HTML]{EDEDED}
\multicolumn{3}{l}{\textit{Spline baselines}} \\
SplineCNN ($K{=}25$) & 88.36k & $97.41 \pm 0.13$ \\
SplineCNN ($K{=}9$) & 55.08k & $96.49 \pm 0.12$ \\
SplineCNN ($K{=}4$) & \redc 44.68k & $94.83 \pm 0.11$ \\
\rowcolor[HTML]{EDEDED}
\multicolumn{3}{l}{\textit{MLP learned-basis control}} \\
MLP basis ($K{=}4$) & \yellowc 45.59k & $97.23 \pm 0.11$ \\
MLP basis ($K{=}16$) & 72.11k & $97.48 \pm 0.07$ \\
MLP basis ($K{=}25$) & 92.00k & $97.32 \pm 0.03$ \\
\rowcolor[HTML]{EDEDED}
\multicolumn{3}{l}{\textit{RBF-GNN with a multivariate basis}} \\
\textbf{RBF-GNN} ($K{=}4$), PCA$+$vp & \orangec 45.26k & $97.79 \pm 0.08$ \\
\textbf{RBF-GNN} ($K{=}9$), PCA$+$vp & 56.38k & \redc $98.00 \pm 0.16$ \\
\textbf{RBF-GNN} ($K{=}9$), spline init & 56.38k & \orangec $97.87 \pm 0.06$ \\
\textbf{RBF-GNN} ($K{=}16$), PCA$+$vp & 71.95k & $97.80 \pm 0.14$ \\
\textbf{RBF-GNN} ($K{=}25$), PCA$+$vp & 91.96k & \yellowc $97.83 \pm 0.11$ \\
\textbf{RBF-GNN} ($K{=}49$), PCA$+$vp & 145.34k & $97.71 \pm 0.17$ \\
\bottomrule
\end{tabular*}
\end{minipage}\hfill
\begin{minipage}[t]{0.49\textwidth}
\vspace{0pt}
\centering\fontsize{7.4pt}{8.2pt}\selectfont
\setlength{\tabcolsep}{3.5pt}
\renewcommand{\arraystretch}{1.00}
\begin{tabular*}{\linewidth}{l@{\extracolsep{\fill}}rc}
\toprule
Method & Params $\downarrow$ & Final $\uparrow$ \\
\midrule
\rowcolor[HTML]{EDEDED}
\multicolumn{3}{l}{\textit{Spline baselines}} \\
SplineCNN ($K{=}243$) & 557.42k & $58.73 \pm 0.26$ \\
SplineCNN ($K{=}32$) & 105.03k & $57.55 \pm 0.34$ \\
\rowcolor[HTML]{EDEDED}
\multicolumn{3}{l}{\textit{MLP learned-basis control}} \\
MLP basis ($K{=}8$) & \orangec 55.39k & \orangec $59.67 \pm 0.32$ \\
MLP basis ($K{=}32$) & 109.96k & $59.02 \pm 0.10$ \\
MLP basis ($K{=}64$) & 182.73k & $57.94 \pm 0.42$ \\
MLP basis ($K{=}128$) & 328.27k & $57.79 \pm 0.07$ \\
\rowcolor[HTML]{EDEDED}
\multicolumn{3}{l}{\textit{RBF-GNN with a multivariate basis}} \\
\textbf{RBF-GNN} ($K{=}4$) & \redc 46.45k & $57.58 \pm 0.23$ \\
\textbf{RBF-GNN} ($K{=}8$) & \yellowc 56.47k & $58.97 \pm 0.16$ \\
\textbf{RBF-GNN} ($K{=}32$) & 116.62k & \redc $59.87 \pm 0.40$ \\
\textbf{RBF-GNN} ($K{=}64$) & 196.81k & \yellowc $59.51 \pm 0.21$ \\
\textbf{RBF-GNN} ($K{=}128$) & 357.19k & $57.83 \pm 0.30$ \\
\bottomrule
\end{tabular*}
\end{minipage}
\end{table}

\paragraph{Initialization.}
Randomly initializing the basis functions works consistently worse. We observe $-0.7$ to $-5.8$ drops in accuracy, see the ablation in our experiments. Since we use small random numbers for initialization the resulting basis functions are approximately flat. First, spatial localization would need to be completely learned from scratch. Second, near-identical basis activations give the different $\Theta_p$ strongly correlated gradients. Therefore, we propose two initializations that allow for spatial localization from the beginning.

\emph{Spline fit.}
Fit each rational function to its B-spline by least squares so the layer starts as approximately SplineConv. We perform a stage-wise optimization to fit our learnable basis functions as follows:
\begin{enumerate}
    \setlength{\itemsep}{2pt}
    \setlength{\topsep}{4pt}
    \setlength{\parsep}{0pt}
    \item We fit the numerator $P$ with a closed-form least squares fit.
    \item Next, we refine both $P$ and $Q$ with Adam (lr $0.01$) for 500 steps.
    \item Last, for further numerical accuracy we use L-BFGS for 100 iterations on float64.
\end{enumerate}
We evaluate squared loss on a fixed grid of points, using 256 for 1D, $32^2$ for 2D and $16^2$ on 3D. We use small values for the denominator $Q$ since the gradient of a constant zero $Q$ would vanish. The fit is cached, so stacked layers share the same fit. The spline fit can be performed whenever the number of basis functions $K = k^D$ for some $k \in \N$.

\emph{PCA of splines.}
When $K \ne k^D$, we cannot fit to the splines. Instead, we fit the $K$ functions to the $K$ leading principal components of the $k^D$ spline basis. In detail, this is done as follows: We sample points on the same grid as above. We collect the values in a matrix, compute its SVD and the $K$ leading left singular vectors $u_1,\ldots,u_K$. We rescale each $u_i$ to unit maximum and fix to a deterministic sign. Last, we fit each $u_i$ with the same procedure as above for the spline fit.

\emph{Variance-preserving gain.}
The initialization of the kernel weights is calibrated for the spline basis. The entries of $\Theta_p$ are drawn i.i.d.\ zero-mean with variance proportional to $1/(K\Cin)$. Thus, the message $\sum_p B_p(\bu)\,\Theta_p\,\bx_j$ along an edge has variance proportional to the basis energy $\sum_p B_p(\bu)^2$. For partition-of-unity basis like the spline the variance remains. However, the PCA-based initialization can result in larger activations and, without correction, to inflated output scale.

Adapting a technique from KAT~\citep{yang2025kat}, we therefore rescale the kernel weights by a suitable value $\sqrt{\alpha}$ after basis initialization. The gain compares the energy of the B-spline basis being replaced with that of the initialized rational basis,
\begin{equation}
\alpha \;=\;
\E_{\bu}\Bigl[\sum_{\mathbf p} N_{\mathbf p}(\bu)^2\Bigr] \Big/\;
\E_{\bu}\Bigl[\sum_p B_p(\bu)^2\Bigr],
\label{eq:vp}
\end{equation}
where $N_{\mathbf p}$ are the $k^D$ tensor-product B-splines, $B_p$ are our $K$ rational basis functions as returned by the spline or PCA fit above, and both expectations are over uniform $\bu$, evaluated on the fit grid. The initial message variance then matches that of SplineConv with the same $K$.

\paragraph{Efficient implementation.}
A naive implementation of \ours{} materialises the Chebyshev feature tensor before contracting it with the coefficients. More efficiently, we provide a Triton kernel that never materializes this intermediate tensor that directly produces only $K$ basis values from the $D$ pseudo-coordinates. The polynomial sums are efficiently computed by a Horner-like scheme. The resulting kernel runs about $15$ times faster than the naive  PyTorch implementation and $3$ times faster than a compiled variant and consumes less memory. Training and inference with the Triton kernel for \ours{} is slightly faster overall than the PyG-implementation of SplineCNN.

\section{Experiments}
\label{sec:experiments}

\setcounter{topnumber}{1}
We evaluate RBF-GNN on eight benchmarks from five tasks, with
pseudo-coordinates of two to six dimensions. In each experiment, RBF-GNN replaces only the SplineConv layers of the network.
The tables compare the following methods.
\begin{description}\setlength{\itemsep}{0pt}\setlength{\parsep}{0pt}
    \item[SplineCNN:] Fixed B-spline basis with $K = k^D$ functions~\citep{fey2018splinecnn}.
    \item[MLP:] Learned basis from a two-layer MLP~\citep{schutt2017schnet,wu2019pointconv}.
    \item[PointNet:] PointNet-style layer on neighbor features and pseudo-coordinates, without a basis~\citep{qi2017pointnet}.
    \item[\ours{} with product basis:] Products of univariate safe Pad\'e functions~\eqref{eq:safepade}.
    \item[\ours{} with multivariate basis:] Non-separable rational functions of all coordinates~\eqref{eq:multivariate}.
\end{description}
Spline init and PCA$+$vp denote the spline fit and the variance-preserving PCA
initialization of Section~\ref{sec:operator}. Results are means and standard
deviations over three seeds unless stated otherwise.

\paragraph{Semantic keypoint matching.}
\label{sec:pascalvoc}
We evaluate semantic keypoint matching in two host pipelines, DGMC~\citep{fey2020dgmc} and the Normalized Matching Transformer (NMT)~\citep{normmatchtrans}, which both use a continuous-kernel graph convolution for geometric feature refinement. In each pipeline we replace only the SplineConv layers and keep the rest of the architecture, the backbone and the training procedure unchanged. Table~\ref{tab:keypoint} reports all configurations for both pipelines on PascalVOC-Keypoints~\citep{everingham2010pascal,bourdev2009poselets} and SPair-71k~\citep{min2019spair}.

In Table~\ref{tab:keypoint}, RBF-GNN achieves the highest accuracy for both pipelines on both datasets. The multivariate basis is best under DGMC on both datasets and, by $0.01$, under NMT on SPair-71k, and the product basis is best under NMT on PascalVOC with $83.78$. The margin over the best spline baseline decreases from $+1.12$ and $+1.62$ under DGMC to $+0.68$ and $+0.01$ under the stronger NMT pipeline, so the accuracy gain depends on the host pipeline. The parameter savings are consistent. In three of the four columns, a rational basis matches or exceeds the widest spline grid with $2.6$ to $4.9\times$ fewer parameters, and under NMT on PascalVOC it is within $0.02$ points of that grid with $2.6\times$ fewer.

\paragraph{Shape correspondence.}
\label{sec:faust}
Dense shape correspondence on FAUST~\citep{bogo2014faust} assigns every vertex
of a held-out human mesh to a template vertex. We follow the SplineCNN
protocol~\citep{fey2018splinecnn}, report exact-match accuracy and compare with
the published SplineCNN result (Table~\ref{tab:faust}). Small multivariate
rational bases achieve the best and most stable results with substantially
fewer parameters than SplineCNN.

\paragraph{Event-based recognition.}
\label{sec:ncaltech101}
\label{sec:ncars}
Both event-camera benchmarks use the seven-layer AEGNN recognition network~\citep{schaefer2022aegnn}, with one graph node per event and space-time pseudo-coordinates, and we replace only its SplineConv layers. We report test accuracy for multiclass object recognition on N-Caltech101~\citep{orchard2015converting}, where we train with augmentation and a validation-based stopping criterion, and for real-world binary car-versus-background recognition on N-Cars~\citep{sironi2018hats}, where we train for 30 epochs. On N-Caltech101, RBF-GNN performs best with max aggregation and also improves substantially over SplineCNN under matched mean aggregation (Table~\ref{tab:ncaltech101}). On N-Cars, RBF-GNN is the most accurate model under both aggregation rules, its mean-aggregation variant outperforms the max-aggregation SplineCNN, and its smaller configuration is more accurate than SplineCNN with fewer parameters (Table~\ref{tab:ncars}).

\paragraph{Superpixel graph classification.}
\label{sec:mnist-superpixels}
\label{sec:cifar10-superpixels}
On MNIST~\citep{lecun1998mnist} superpixels, we follow the SplineCNN protocol~\citep{fey2018splinecnn} and use two continuous-kernel convolution layers on 2-D Cartesian pseudo-coordinates. On CIFAR-10~\citep{krizhevsky2009cifar} superpixels from GNNBenchmarkDataset~\citep{dwivedi2023benchmarking}, we use a 5-D bilateral pseudo-coordinate that concatenates the relative 2-D centroid position with the RGB features. We report test accuracy on both datasets. On MNIST, RBF-GNN achieves the three highest accuracies and improves over SplineCNN at both matched and reduced basis sizes (Table~\ref{tab:mnist-superpixels}). Wider bases do not help: $K{=}16$ and $25$ stay within noise of $K{=}9$, and $K{=}49$, initialized on a $k{=}7$ spline grid, is slightly worse. On CIFAR-10, \ours{} attains the best accuracy at $K{=}32$ while using fewer parameters than the wider spline grid (Table~\ref{tab:cifar10-bilateral}).

\paragraph{Part segmentation.}
\label{sec:shapenet-part}
For point-cloud part segmentation on ShapeNet-Part~\citep{yi2016scalable}, we
use a fixed subset of 4,000 training and 1,333 test shapes. Relative positions
and surface normals define 6-D pseudo-coordinates on a 16-nearest-neighbor
graph, and we report instance-average mIoU.
RBF-GNN nearly matches the large
spline grid with an order-of-magnitude smaller model and outperforms the
equal-width MLP control (Table~\ref{tab:shapenet-part}). From $K{=}32$ on, accuracy
plateaus within noise of the $729$-function spline grid ($p \geq 0.27$) at $2$ to
$12\times$ fewer parameters, with the MLP control below it at every width from $K{=}32$ on. At the smallest width
the MLP control is more accurate, so the advantage of RBF-GNN on this benchmark
is parameter efficiency at larger widths, not a gain at every width.

\paragraph{Ablation study.}
\label{sec:ablation}
\textit{Basis size.} For the PCA$+$vp rows of Table~\ref{tab:keypoint}, accuracy rises from $K{=}1$ to $K{=}4$ in all four columns, peaks at $K{=}6$ for DGMC and changes little beyond it for NMT. At matched $K$, the advantage over SplineCNN is concentrated at small bases. It ranges from $+0.42$ to $+5.52$ for $K \in \{4,9\}$, and at $K{=}25$ it is negligible, between $-0.02$ and $+0.16$, except under NMT on PascalVOC with $+0.68$. On FAUST, the multivariate basis degrades sharply, with high variance across seeds, at $K{=}16$ and $K{=}27$ (Table~\ref{tab:faust}), which points to a fragile PCA initialization rather than insufficient capacity. At 5-D and 6-D, in contrast, RBF-GNN improves up to $K{=}32$ and then saturates on ShapeNet-Part and declines on CIFAR-10 (Tables~\ref{tab:cifar10-bilateral} and~\ref{tab:shapenet-part}).

\textit{Basis form and polynomial degree.} At $K{=}9$, the multivariate basis with spline init outperforms the product basis in all four columns of Table~\ref{tab:keypoint}, but with higher polynomial degrees, $(8,6)$ instead of $(5,4)$. In the degree control, where both forms use $(5,4)$ at $K{=}25$, they differ by at most $0.43$ points. On FAUST, the product basis remains stable at $K{=}27$, whereas the multivariate basis degrades sharply (Table~\ref{tab:faust}).

\textit{Initialization.} The tables compare the two initializations only at $K{=}9$, where spline init outperforms PCA$+$vp by $0.18$ to $0.52$ points in all four columns of Table~\ref{tab:keypoint} and is $0.13$ points worse on MNIST (Table~\ref{tab:mnist-superpixels}). The two therefore perform similarly, and PCA$+$vp is required whenever $K \neq k^D$.

\textit{Aggregation.} On FAUST, SplineCNN is most accurate with mean aggregation and unstable with add aggregation unless gradients are clipped, whereas RBF-GNN at $K{=}4$ is more accurate with add than with mean (Table~\ref{tab:faust}). On both event-camera benchmarks, max aggregation outperforms mean aggregation for SplineCNN and RBF-GNN alike, and since RBF-GNN is more accurate under both rules, its gain is not explained by the choice of aggregation alone (Tables~\ref{tab:ncaltech101} and~\ref{tab:ncars}).

\textit{Structured and learned bases.} In Table~\ref{tab:keypoint}, the best rational basis in each column exceeds the best MLP control by $+1.36$, $+3.46$, $+1.66$ and $+1.06$ points. Unlike the margin over SplineCNN, this margin does not decrease systematically from DGMC to NMT, which indicates that the rational form itself matters and not only the fact that the basis is learned. On MNIST, the rational parametrization outperforms the equal-width MLP control at every width, whereas on CIFAR-10 and ShapeNet-Part the MLP control is more accurate only at the smallest shared width. On both event-camera benchmarks, the PointNet control uses fewer parameters than \ours{} but is less accurate.

\section{Conclusion}
We have proposed \ours{}, a drop-in replacement for splines in SplineCNN \citep{fey2018splinecnn}.
Experiments show consistent improvements in performance across a wide range of methods where SplineCNN has been used before.
We argue that hand-designed activation functions as used in SplineCNN are worse than learnable ones as used in \ours{}, which encode spatial relationships more flexibly but still encode an important inductive bias in contrast to a pure MLP basis.

\FloatBarrier
\setcounter{topnumber}{2}

\newpage
\subsection*{AI use statement}

In this work, we used generative AI tools to assist in implementing, refactoring and debugging the experimental code. We have not used generative AI tools to generate synthetic data, to formulate mathematical claims or to assist with translation, and the writing of proofs as well as qualitative and thematic data analysis are not applicable to this work. Additionally, we used generative AI tools to polish the language and improve the readability of the manuscript. We have reviewed all AI-assisted work. AI-assisted code was verified and tested for correctness by the authors, and the reported results were checked against the logs of the corresponding runs. The authors take full responsibility for the integrity, accuracy, and entire content of the
submission.

\subsection*{Ethics statement}

All authors have read and adhere to the ICLR Code of Ethics. This work proposes a basis for continuous-kernel graph convolutions and evaluates it on established benchmarks that are available to the research community. We collected no new data and conducted no experiments with human participants. Some of the benchmarks contain images or three-dimensional scans of people. We used them only as distributed by their creators and in accordance with their licences and terms of use, several of which restrict use to non-commercial research, and we made no attempt to identify any individual. The contribution is methodological and not tied to a particular application, so we do not foresee harmful uses beyond those that apply to graph learning methods in general.

\subsection*{Reproducibility statement}

Our code will be released publicly under an open-source licence upon acceptance. Section~\ref{sec:operator} specifies the rational basis, both initialisation procedures and the variance-preserving rescaling, and Section~\ref{sec:experiments} states the task, the metric and the number of random seeds for every benchmark. All reported results are means and standard deviations over independent runs with different seeds. Apart from the replaced basis, every host architecture keeps its published configuration unless Section~\ref{sec:experiments} states otherwise.

\bibliographystyle{iclr2027_conference}
\enlargethispage{2\baselineskip}
\bibliography{rational_basis}

\end{document}